\documentclass[10pt,twocolumn,letterpaper]{article}

\usepackage{cvpr}              

\usepackage{graphicx}
\usepackage{amsmath}
\usepackage{amssymb}
\usepackage{booktabs}
\usepackage{multicol}
\usepackage{multirow}

\usepackage[pagebackref,breaklinks,colorlinks]{hyperref}

\usepackage[capitalize]{cleveref}
\crefname{section}{Sec.}{Secs.}
\Crefname{section}{Section}{Sections}
\Crefname{table}{Table}{Tables}
\crefname{table}{Tab.}{Tabs.}

\def\cvprPaperID{*****} 
\def\confName{CVPR}
\def\confYear{2022}

\begin{document}

\title{Exploiting Context Information for Generic Event Boundary Captioning}

\author{Jinrui Zhang$^1$, Teng Wang$^{12}$, Feng Zheng$^1$, Ran Cheng$^1$, Ping Luo$^2$\\
$^1$Southern University of Science and Technology, $^2$The University of Hong Kong\\
{\tt\small \{zhangjr2018, wangt2020\}@mail.sustech.edu.cn,}
{\tt\small \{zhengf, chengr\}@sustech.edu.cn,}
{\tt\small  pluo@cs.hku.hk}
}

\maketitle

\begin{abstract}
Generic Event Boundary Captioning (GEBC) aims to generate three sentences describing the status change for a given time boundary. Previous methods only process the information of a single boundary at a time, which lacks utilization of video context information. To tackle this issue, we design a model that directly takes the whole video as input and generates captions for all boundaries parallelly. The model could learn the context information for each time boundary by modeling the boundary-boundary interactions. Experiments demonstrate the effectiveness of context information. The proposed method achieved a 72.84 score on the test set, and we reached the $2^{nd}$ place in this challenge. Our code is available at: \url{https://github.com/zjr2000/Context-GEBC}
\end{abstract}
\section{Introduction}
The goal of GEBC is to generate three captions that describe the given event boundary: one for the subject of the event boundary and the other two for the status before and after the boundary. The key challenge of this task is understanding the event that happens around the time boundary. Previous methods~\cite{wang2022geb+} only rely on a single clip around the time boundary for captioning, which means they could not get any information about other events. This limits their ability to exploit the event-event interaction. We argue that this kind of context information plays a vital role in understanding the events. There are two key observations: 1. The event boundaries' subjects of the same video are highly likely to be related. Some boundaries even share the same subject. 2. There are some interactions among events. An event may be the cause of another event. The former could help the model to better capture the subject of event boundaries. The latter could help the model to understand the cause of the event, which could benefit the event understanding.

To utilize the context information of the whole video, we design a parallel captioning architecture. The model takes the whole video and all event boundaries as the input and generates a caption for all boundaries in parallel. We use the self-attention mechanism~\cite{vaswani2017attention} to let each boundary interact with all video frames. Benefit from the excellent pairwise modeling ability of self-attention. Each boundary could not only attend to the frames around the boundary but also capture the key information from other events.

In summary, we propose that the video context information could benefit the GEBC task. We construct a parallel captioning model to utilize context information. The experiment results demonstrate the effectiveness of the proposed method.

\section{Method}
\begin{figure*}
\vspace{-2em}
  \includegraphics[width=1.0\textwidth]{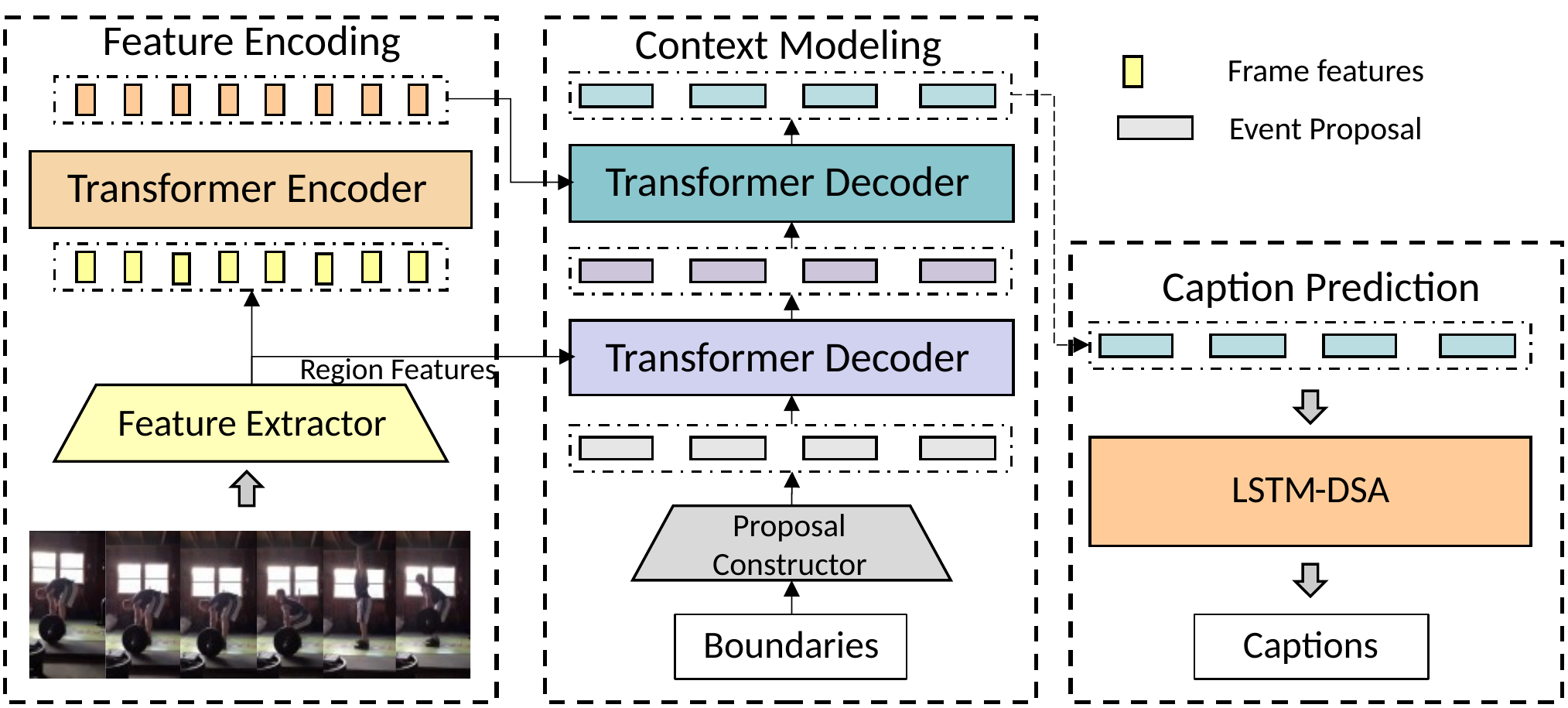}
  \caption{Illustration of the proposed model. Our model contains a transformer encoder for feature encoding, two transformer decoder for context modeling and a caption head for caption prediction. We train three separate models for subject captioning and two types of status captioning.}
  \label{fig:model}
\end{figure*}

Formally, GEBC can be formulated as follows. Given a video with $\bf{T}$ frames and $\bf{N}$ event boundaries, denoted by $\{t^1, t^2, ..., t^N\}$. The model need to generate three captions for each event boundary: $\bf{S}_{bef}$, $\bf{S}_{aft}$ and $\bf{S}_{sub}$, which descibe the status before the boundary, the status after the boundary and the subject, respectively. As shown in Fig~\ref{fig:model}, our model contains a transformer encoder for feature encoding, two transformer decoder for context modeling and a caption head for caption prediction. 

\subsection{Feature Encoding}
To better exploit spatial and temporal information from the video content, we adopt multiple pre-trained models to extract visual features. We use Omnivore~\cite{girdhar2022omnivore} and CLIP~\cite{radford2021learning} to extract frame features. Specifically, we uniformly sample a sequence of frames with a stride of $\bf{m}$ frames and then extract features for each sampled frame. To facilitate batch processing, we perform an interpolation strategy to resize the temporal dimension of the features to a fixed number $\bf{L}$. The two kinds of frame features are directly concatenated as the final frame-level features. The frame-level features are fed into a transformer encoder to further exploit frame-frame interactions. The processed frame-level features are denoted by $\bf{F} \in \mathbb{R}^{\bf{L} \times d}$, where $\bf{d}$ is the hidden dimension of transformer. We use VinVL~\cite{zhang2021vinvl} to generate region-level features. Specifically, we first select a frame for each clip between two boundaries and then extract the features. Note that the selected frame for the clip is the one nearest to the center timestamp. To align the dimension, the region features are projected to $\mathbb{R}^{d}$, thus, the final region-level features are denoted by $\bf{O} \in \mathbb{R}^{(\bf{N} + 1) \times N_o \times d}$, where $N_o$ denotes the number of objects per frame.

\subsection{Context Modeling}
 To model the relationship between event boundaries and visual features. We first construct event proposals from event boundaries. Because three different kinds of captions focus on different time periods relative to the boundary, \textit{e.g.}, $\bf{S}_{sub}$ focus on the time periods before the given boundary, we construct three different proposals for three caption generation. Formally, for the $i$-th boundary $t^i$, three time boxes $(t^{i-1}, t^{i+1})$, $(t^{i-1}, t^{i})$ and $(t^{i}, t^{i+1})$ are utilized for the proposal construction of $\bf{S}_{sub}$, $\bf{S}_{bef}$ and $\bf{S}_{aft}$, respectively. Each time boxes are first normalized to $[0, 1]$, and then processed by inverse sigmoid function. Position embedding and layer normalization are performed to convert them to vectors. All time boxes are projected to $\mathbb{R}^{d}$, thus, the event proposals are denoted by $\bf{E} \in \mathbb{R}^{N \times d}$. Note that, we train three seperate models for $\bf{S}_{sub}$, $\bf{S}_{bef}$ and $\bf{S}_{aft}$, so there are only $\bf{N}$ event proposals for each model.

The event proposals are first used to query region-level features, which aims to facilitate the model attending to the key subjects in the current timestamp. After that, the frame-level features are queried to gather the context information that is relevant to the current event proposal. Each event boundary may capture the cause of the state change by this operation, which helps the model understand the event from another perspective. The outputs of the transformer decoder are the features of event boundaries. 

\subsection{Caption Prediction}
The output visual features for $\bf{N}$ event boundaries would be fed into a caption head to generate captions. We follows~\cite{wang2021end} to use LSTM-DSA as caption head to perform auto-regressive sentence generation. The model will generate one word on each iteration until a special token $<$end$>$ is generated or meet the maximum length $\bf{M}$. We train the model with the cross-entropy between the predicted word probability and the ground truth.
\section{Experiments}

\subsection{Implementation Details}

\begin{table*}[]
\caption{Performance on the validation set.}
\renewcommand\arraystretch{1.3}
\centering
        \makeatletter\def\@captype{table}\makeatother
        \setlength{\tabcolsep}{1.3 mm}{
\begin{tabular}{l|cccc|cccc}
\toprule
      \multirow{2}{*}{Model} &\multicolumn{4}{c|}{CIDEr} & \multicolumn{4}{c}{ROUGE\_L} \\
&  Average   & Subject   & Before   & After &  Average   & Subject   & Before   & After \\ 
\midrule
Official baseline (CLIP feature) & 87.72 & 148.50 & 61.87 & 52.78 & 31.02 & 49.27 & 22.28 & 21.50 \\ 
CLIP features only & 117.53 & 176.91 & 93.93 & 81.76 & 35.95 & 52.33 & 28.79 & 26.74 \\
+ Omnivore features & 126.51 & 192.06 & 98.23 & 89.23 & 36.94 & 53.19 & 29.32 & 28.33 \\
+ Query region features & 140.95 & 228.79 & 102.74 & 91.31 & 38.04 & 55.52 & 29.96 & 28.65 \\
+ Reinforcement learning & \textbf{149.48} & \textbf{235.25} & \textbf{109.00} & \textbf{104.2} & \textbf{39.86} & \textbf{56.48} & \textbf{31.50} & \textbf{31.60} \\

\bottomrule
\end{tabular}}
\label{tab:eval}
\end{table*}
The code is implemented based on HugginFace Transformers~\cite{wolf-etal-2020-transformers} and PDVC~\cite{wang2021end}. We use the full train set to train the model. The object number per frame $\bf{N}_o$ is set to 50. Zero padding is added if the object number is less than $\bf{N}_o$. We take top $N_{o}$ objects sorted by their confidence if one frame has more than $\bf{N}_o$ objects. The frame sample interval $\bf{m}$ is set to 8 and 16 for CLIP and Omnivore, respectively. The fixed-length $\bf{L}$ is set to 100. The maximum sentence length is set to $30$ for all three kinds of captions. Both transformer encoder and transformer decoder are implemented based on deformable transformer~\cite{zhu2020deformable}. The transformer decoder for region feature modeling contains one layer, and the other two transformer architectures contain two layers. The hidden dimension $\bf{d}$ of the transformer is set to 512. We train the model with the AdamW optimizer~\cite{loshchilov2017decoupled}. The weight decay and mini-batch size are set to $0.0001$ and $8$, respectively. The initial learning rate is set to $5 \times e^{-5}$. The learning rate decays from $8$-th epoch and by a factor of 0.5 per 3 epochs.



\subsection{Evaluation Results}
Our model achieves 72.84 scores on the test set of this challenge, with a relative improvement of $78.53\%$~($\frac{72.84-40.80}{40.80}$) over the official baseline method (ActBERT-Revised~\cite{wang2022geb+}). The biggest difference between baseline and our method is that our model exploits context information from the whole video content. The baseline method only can see the information about a single boundary, which limits their ability to get more semantic information. This could demonstrate the importance of context information to the GEBC task.

Table~\ref{tab:eval} shows several techniques that boost generic event boundary captioning systems. We show their evaluation results on the validation set in terms of CIDEr and ROUGE\_L. The combination of the CLIP feature and Omnivore feature could clearly improve $5\%$ performance (117.53 $\longrightarrow$ 126.51) compared with a single feature. To facilitate the model attending to the key objects in videos, we further introduce region-level features. We use event proposals to query region features. From the table, we can see that this technique clearly yields a performance gain. It is worth noting that the score of ``subject" gains the most obvious improvement (around $19\%$), which shows that region-level features help to capture the subjects of the event. Then we use the CIDEr score as a reward to train the model with a reinforcement learning strategy, which gives a considerable performance on all evaluation metrics.
\section{Conclusion}
We propose a context modeling method for event boundary captioning. Previous methods only process the information of single event boundary, which lack the perception of the global video contexts. To tackle this issue, we propose a model that takes the whole video as input and generates captions of all event boundaries in parallel. Experiment results demonstrate the importance of context information to the GEBC task. The proposed method achieves a $72.84$ score on the test set, which significantly outperforms the official baseline methods.


{\small
\bibliographystyle{ieee_fullname}
\bibliography{egbib}
}

\end{document}